\documentclass[11pt]{article}
\usepackage[a4paper]{geometry}
\usepackage{clic2017}
\usepackage{times}
\usepackage{url}
\usepackage{latexsym}
\usepackage{tabularx}
\usepackage{multirow}
\usepackage{xcolor}

\title{Concept Tagging for Natural Language Understanding:\\Two Decadelong Algorithm Development}

\author{Jacopo Gobbi \\
  University of Trento \\
  Trento, Italy \\
  {\small\tt jacopo.gobbi@studenti.unitn.it} \\\And
  Evgeny A. Stepanov \\
  VUI, Inc. \\
  Trento, Italy \\
  {\small\tt eas@vui.com} \\\And
  Giuseppe Riccardi\\
  University of Trento \\
  Trento, Italy \\
  {\small\tt giuseppe.riccardi@unitn.it} \\}

\date{}

\begin{document}
\maketitle

\newcolumntype{R}{>{\raggedleft\arraybackslash}X}

\begin{abstract}
  \textbf{English.} Concept tagging is a type of structured learning needed for natural language understanding (NLU) systems. In this task, meaning labels from a domain ontology are assigned to word sequences. In this paper, we review the algorithms developed over the last twenty five years. We perform a comparative evaluation of generative, discriminative and deep learning methods on two public datasets. We report on the statistical variability performance measurements. The third contribution is the release of a repository of the algorithms, datasets and recipes for NLU evaluation.
\end{abstract}

\begin{abstract-alt}
 \textrm{\bf{Italiano.}} L'annotazione automatica dei concetti \`{e} un tipo di apprendimento strutturato necessario per i sistemi di comprensione del linguaggio naturale (NLU). In questo processo le etichette di un'ontologia di dominio sono assegnate a sequenze di parole. In questo articolo esaminiamo gli algoritmi sviluppati negli ultimi venticinque anni. Eseguiamo una valutazione comparativa dei metodi di apprendimento generativo, discriminatorio e approfondito su due set di dati pubblici. Il secondo contributo \'{e} un'analisi della variabilit\'{a} delle misure di valutazione. Il terzo contributo  \`{e} il rilascio di un archivio degli algoritmi, dei sets di dati e delle ricette per la valutazione dell'NLU.
\end{abstract-alt}

\section{Introduction} 
The NLU component of a conversational system requires an automatic extraction of concept tags, dialogue acts, domain labels and entities. In this paper we describe and review the algorithm development of the concept tagging (a.k.a. slot filling or entity extraction) task. It aims at computing a sequence of concept units, $C=c_1..c_M$, from a sequence of words in natural language, $W=w_1..w_N$. The task can be seen as a structured learning problem where words are the input and concepts are the output labels. 
In other words, the objective is to map a sentence (utterance)  ``\textit{I want to go from Boston to Atlanta on Monday}'' to the sequence of domain labels``{\small\texttt{null null null null null fromloc.city null toloc.city null depart\_date.day\_name}}'', that would allow to identify, for instance that \textit{Boston} is a \textit{departure city} . Difficulties may arise from different factors, such as the variable token span of concepts, the long-distance word dependencies, a large and ever changing vocabulary, or subtle semantic implications that might be hard to capture at a surface level or without some prior context knowledge.

Since the early nineties \cite{PIERACCINI1992283}, the task has been designed as a core component of the natural language understanding process in domain-limited conversational systems. Over the years, algorithms have been developed for generative, discriminative and, more recently, for deep learning frameworks. In this paper, we provide a comprehensive review of the algorithms, their parameters and their respective state-of-the-art performances. We discuss the relative advantages and differences amongst algorithms in terms of performances and statistical variability and the optimal parameter settings. Last but not least, we have designed and provided a repository of the data, algorithms, implementations and parameter settings on two public datasets. The GitHub repository\footnote{\scriptsize www.github.com/fruttasecca/concept-tagging-with-neural-networks} is intended as a reference both for practitioners and for algorithm development researchers.

 

With the conversational AI gaining popularity, the area of NLU is too vast to mention all relevant or even recent studies. Moreover the objective of this paper is to benchmark a important subtask of NLU, concept tagging used by advanced conversational systems. We benchmark generative, discriminative and deep learning approaches to NLU, the work is in-line with the works of \cite{conf/interspeech/RaymondR07,mesnil2015using,bechet_raymond_IS2018}.
Unlike mentioned previous comparative performance analysis, in this paper, we benchmark deep learning architectures, and compare them to a generative and traditional discriminative algorithms. To the best of our knowledge, this is the first comprehensive comparison of concept tagging algorithms at this scale on public datasets and shared algorithm implementations ( and their parameter settings).

\section{Algorithms}
Among the algorithms considered for benchmarking, we include a representative from the generative class, the weighted finite state transducers (WFSTs); and two discriminative algorithms: Support Vector Machines (SVMs), Conditional Random Fields (CRFs), and a set of base neural networks architectures and their combinations.

\indent \textbf{Weighted Finite State Transducers}\footnote{We use OpenFST (http://www.openfst.org) and OpenGRM (http://www.opengrm.org) libraries.} cast concept tagging as a translation problem from words to concepts \cite{conf/interspeech/RaymondR07}, and usually consist of two components.
The first component transduces words to concepts based on a score that can be either induced from data or manually designed; the second component is a stochastic conceptual language model, which re-scores concept sequences. The two components are composed to perform sequence-to-sequence translation and infer the best sequence using Viterbi algorithm.

\indent \textbf{Support Vector Machines (SVM)} are used within
Yamcha tool \cite{Kudo:2001:CSV:1073336.1073361} that performs sequence labeling using forward and backward moving classifiers. Automatic labels assigned to preceding tokens are used as dynamic features for the current token's label decision.

\indent \textbf {Conditional Random Fields (CRF)}\footnote{We use CRFSUITE \cite{Okazaki:2007} implementation of CRFs in out experiments.} \cite{Lafferty:2001:CRF:645530.655813} is a discriminative model based on a dependency graph $G$ and a set of features. Each feature $f_k$ has an associated weight $\lambda_k$. Features are generally hand-crafted and their weights are learned from the training data.
Additionally, we experiment with word embeddings as additional features for CRFs (CRF+EMB).

\indent \textbf {Recurrent Neural Networks (RNN)}.
The first neural network architecture\footnote{All neural architectures are implemented within the PyTorch framework (https://pytorch.org)} we have considered is an Elman RNN \cite{Elman90findingstructure,beyli2012CaseSF}. In RNN, a hidden state depends on the current input and the previous hidden state. The output (label), on the other hand, depends on the new hidden state. 

\indent \textbf {Long-Short Term Memory (LSTM)}
RNNs \cite{Hochreiter:1997:LSM:1246443.1246450} try to tackle the vanishing gradient problem by introducing a more complex mechanisms to address information propagation and deletion, with the cost of a more complex model with more parameters to train  due to the system of gates it uses.
The memory of the model is represented by the cell state and the hidden state, which also represents the output for the current token.
We experimented with a simple LSTM, an LSTM which receives as input the word embedding concatenated with character embeddings obtained through a convolutional layer \cite{DBLP:journals/corr/JozefowiczVSSW16} (LSTM-CHAR-REP), and an LSTM with pre-trained embeddings and dynamic embeddings learned from training data (LSTM-2CH). In LSTM-2CH two separate LSTM modules run in parallel and their outputs are concatenated for each word. Similar to the rest of the deep learning models, the output is then fed to a fully connected layer to map every token to the concept tag space.

\indent \textbf {Gated Recurrent Units (GRU)} \cite{DBLP:journals/corr/ChoMGBSB14} use a reset and an update gate, which are two vectors of weights that decide what information is deleted (or re-scaled) from the current hidden state and how it will contribute to the new hidden state, which will also be the output for the current input.
Compared to the LSTM model this allows to train fewer parameters, but introduces a constraint on memory, since it is also used as an output.

\indent \textbf {Convolutional Neural Networks (CONV)}
\cite{Majumder:2017:DLD:3080885.3080974,kim2014convolutional} consider each sentence as a matrix of shape (\# words in sentence, size of embedding) for convolution using kernels of different sizes to pass over the input sequence token-by-token, bigram by bigram and trigram by trigram. The result of convolution is used as a starting hidden memory for a GRU RNN.
GRU RNN is used on embedded tokens and starts with the information on the sequence at a global level.

\indent \textbf {FC-INIT} is similar to CONV. The difference is in the pre-elaboration of the hidden state, which is done by fully connected layers elaborating on the whole sequence. 

\indent \textbf {ENCODER} architecture \cite{DBLP:journals/corr/ChoMGBSB14} casts the problem as a sequence-to-sequence translation and consists of two GRU RNNs. Encoder, the first GRU RNN, encodes the input sequence to a fixed vector (the hidden state). Decoder, another GRU RNN, uses the output of the encoder as a starting hidden state. At each step, the decoder receives the label predicted at the previous step as an input, starting with a special token.

\indent \textbf {ATTENTION} architecture is similar to ENCODER with the addition of an attention mechanism \cite{DBLP:journals/corr/BahdanauCB14} on the outputs of the encoder. This allows the network to focus on a specific parts of the input sequence. The attention weights are computed with a single fully connected layer that receives as input the embedding of the current word concatenated to the last hidden state.

\indent \textbf {LSTM-CRF}
 \cite{recurrent-conditional-random-field-for-language-understanding,Zheng:2015:CRF:2919332.2919659} is
an architecture where the LSTM provides class scores for each token, and the Viterbi algorithm decides on the labels of the sequence at a global level using bigrams and transition probabilities that are trained with the rest of the parameters.
We also experimented with a variant that considers character level information (LSTM-CRF-CHAR-REP).

\section{Corpora}
The evaluation of algorithms is performed on two datasets.
The Air Travel Information System (\textbf{ATIS}) dataset consists of sentences from users querying for information about flights, departure dates, arrivals, etc.
The training set consists of 4,978 sentences, while there are 893 sentences that constitute the test set. The average length of a sentence is around 11 tokens, and there are a total of 127 unique tags (with IOB prefixes). Moreover, the large majority of tokens missing an embedding are either numbers or airport/basis/aircraft codes. The training set has a total of 18 types missing an embedding, and the test set has 9. 

The second corpus (\textbf{MOVIES}) was produced from NL-SPARQL \cite{chen2014deriving} corpus semi-automatically aligning SPARQL query values to utterance tokens. The dataset follows the split of the original corpus having 3,338 sentences (with 1,728 unique tokens) and 1,084 sentences (with 1,039 tokens) in the training and test sets, respectively. The average length of a sentence is  6.50 and the OOV rate is 0.24. There are 43 concept tags in the dataset.
Given the Google embeddings, once we consider every number as a class {\em  number}, we obtain 66 token types without an embedding for the training set and 26 for the test set.

\begin{table}[t]
\centering
\begin{small}
\setlength\tabcolsep{3pt}
\begin{tabularx}{\columnwidth}{|l|X|r|r|}
\hline
\textbf{Model} 
& \multicolumn{1}{X|}{\textbf{Parameters}}
& \multicolumn{1}{c|}{\textbf{\# Params}}
& \multicolumn{1}{c|}{\textbf{$F_1$}}\\
\hline
\multirow{2}{*}{\textit{WFST}}

&  {\tiny order 4, kneser ney }  & {\tiny (7907 states, 842178 arcs) } & 82.96 \\
& {\tiny order 4, kneser ney }  & {\tiny (4124 states, 76000 arcs)} & 93.08 \\
\hline
\multirow{2}{*}{\textit{SVM}}
& {\tiny (−4, 4) window of tokens, (-1, 0) of POS tag and prefix. Postfix and lemma of current word. Previous two labels.}
&10364
& 83.74 \\
& {\tiny (−6, 4) window of tokens, (-1, 0) of prefix and postfix. Previous two labels .}
&16361
& 92.91 \\
\hline

\multirow{2}{*}{\textit{CRF}}
& {\tiny (−4, 4) window of token, (-1, 0) of POS tag and prefix. Postfix and lemma of current word.  Previous + current word conjunction, current + next word conjunction. Bigram model.}
& 1,200K& 83.80 \\
& {\tiny (−6, 4) window of tokens, (-1, 0) of  prefix. Postfix of current word. Previous + current word conjunction. Bigram model.}
& 2,201K& 93.98 \\
\hline
\multirow{2}{*}{\textit{CRF+EMB}}
& {\tiny all above +  (−4, 4) word embs + current token char embeddings}
& 1,390K& 85.85 \\
& {\tiny all above +  (−6, 4) word embs + current token char embeddings} & 3,185K& 94.00 \\
\hline
\end{tabularx}
\caption{Performance of the WFST, SVM and CRF (with and without embeddings) algorithms. For each algorithm we report F$_1$ score for the MOVIES (top row) and  ATIS (bottom row) datasets. }
\label{tbl:results_ml}
\end{small}
\end{table}

\begin{table*}[t]
\centering
\begin{small}
\setlength\tabcolsep{3pt}
\begin{tabularx}{\textwidth}{|l|R|R|R|R|R|R|R|R|R|R|}
\hline
\textbf{Model} 
& \multicolumn{1}{X|}{\textbf{hidden}} 
& \multicolumn{1}{X|}{\textbf{epochs}}
& \multicolumn{1}{X|}{\textbf{batch size}}
& \multicolumn{1}{c|}{\textbf{lr}}
& \multicolumn{1}{X|}{\textbf{drop rate}}
& \multicolumn{1}{X|}{\textbf{emb norm}}
& \multicolumn{1}{X|}{\textbf{\# of params}}
& \multicolumn{1}{X|}{\textbf{min $F_1$}}
& \multicolumn{1}{X|}{\textbf{avg $F_1$}}
& \multicolumn{1}{X|}{\textbf{best $F_1$}}\\
\hline
\multirow{2}{*}{\textit{RNN}} 
& 200 & 15 & 50 & 0.001 & 0.30 &  4 & 1,264K & 81.00 & 82.55 & 83.96 \\
& 400 & 10 & 50 & 0.001 & 0.25 &  2 &   580K & 91.80 & 93.79 & 95.03 \\
\hline
\multirow{2}{*}{\textit{LSTM}} 
& 200 & 15 & 20 & 0.001 & 0.70 &  6 & 1,505K & 82.67 & 83.76 & 84.57 \\
& 200 & 15 & 10 & 0.001 & 0.50 & 8 & 675K &  87.82 &  94.53 &  95.36\\
\hline
\multirow{2}{*}{\textit{LSTM-CHAR-REP}} 
& 400 & 20 & 20 & 0.001 & 0.70 &  4 & 2,085K & 82.00 & 84.28 & 85.41 \\
& 400 & 15 & 10 & 0.001 & 0.50 &  6 & 1,272K & 81.00 & 94.19 & 95.39 \\
\hline
\multirow{2}{*}{\textit{LSTM-2CH}} 
& 200 & 20 & 15 & 0.001 & 0.30 &  8 & 1,310K & 81.22 & 82.68 & 83.76 \\
& 400 & 10 &100 & 0.010 & 0.70 &  6 & 1,022K & 93.10 & 94.61 & 95.38 \\
\hline
\multirow{2}{*}{\textit{GRU}} 
& 200 & 20 & 20 & 0.001 & 0.50 &  4 & 1,424K & 76.56 & 84.29 & 85.47 \\
& 100 & 15 & 10 & 0.005 & 0.50 & 10 &   446K & 91.53 & 94.28 & 95.28 \\
\hline
\multirow{2}{*}{\textit{CONV}} 
& 200 & 20 & 20 & 0.001 & 0.50 &  4 & 2,646K & 84.05 & 85.02 & 86.17 \\
& 100 & 15 & 10 & 0.005 & 0.00 &  2 &   625K & 91.51 & 94.22 & 95.38 \\
\hline
\multirow{2}{*}{\textit{FC-INIT}} 
& 100 & 30 & 20 & 0.001 & 0.30 &  4 & 2,805K & 82.22 & 83.93 & 84.95 \\
& 400 & 15 & 50 & 0.010 & 0.25 &  4 & 7,144K & 87.39 & 94.67 & 95.39 \\
\hline
\multirow{2}{*}{\textit{ENCODER}} 
& 200 & 30 & 20 & 0.001 & 0.70 &  4 & 1,559K & 71.25 & 76.39 & 79.00 \\
& 200 & 25 &  5 & 0.001 & 0.70 &  6 &   730K & 70.01 & 78.16 & 80.85 \\
\hline
\multirow{2}{*}{\textit{ATTENTION}} 
& 200 & 15 & 20 & 0.001 & 0.30 &  4 & 1,712K & 71.86 & 79.77 & 82.67 \\
& 200 & 25 &  5 & 0.001 & 0.25 & 10 &   894K & 92.47 & 94.09 & 94.98 \\
\hline
\multirow{2}{*}{\textit{LSTM-CRF}} 
& 200 & 10 &  1 & 0.001 & 0.70 &  6 & 1,507K & 84.75 & \textbf{86.11} & 87.47 \\
& 400 & 15 & 10 & 0.001 & 0.50 &  6 & 1,200K & 94.39 & 94.72 & 95.01 \\
\hline
\multirow{2}{*}{\textit{LSTM-CRF-CHAR-REP}} 
& 200 & 15 & 1 & 0.001 & 0.70 & 8 & 1,555K & 85.07 & 86.08 & 87.05\\
& 200 & 20 & 5 & 0.001 & 0.50 &  4 &   740K & 94.45 & \textbf{94.91} & 95.12 \\
\hline
\end{tabularx}
\end{small}
\caption{ All models are bidirectional and have been trained with unfrozen Google embeddings, except for CONV and LSTM-2CH. Min, average and best F$_1$ scores are obtained training the same model with the same hyperparameters, but different parameter initializations. Averages are from 50 runs for MOVIES and 25 for ATIS. For each architecture, the first row reports F$_1$-score for the MOVIES dataset and the second for ATIS.
Hyperparameter search has been done randomly over ranges of values taken from published work.
The number of parameters refers to the network parameters plus the embeddings, when those are unfrozen.
Given a hidden layer size $X$ reported in \textbf{hidden} column, each component in the bidirectional architecture would have a hidden layer size of $X/2$. Similarly, each of the two LSTM components in the LSTM-2CH model would have $X/2$ as an hidden layer size; and each bidirectional component would thus have a hidden layer size equal to $X/4$.} 
\label{tbl:results_nn}
\end{table*}

\section{Performance Analysis }
One of our first observations is the fact that models such as WFST, SVM and CRF yield competitive results with simple setups and few hyperparameters to be tuned. The training of our deep learning models and the search of their hyperparameters would have been unfeasible without dedicated hardware, while it took a fraction of the effort for WFST, SVM and CRF. Moreover, adding word embeddings as features to the CRF allowed it to outperform most of the deep neural networks.


We attribute this to two factors: (1) since these models, unlike neural networks, 
do not learn feature representation from data,
they are simpler and faster to train; and, most importantly, (2) these models usually perform global optimization over the label sequence, while neural networks usually do not.
Augmenting neural networks with CRF is not expensive in terms of parameters. Having a CRF component on top of an LSTM increments the number of parameters up to the square of the tag-set size (about 2,500 for the MOVIES dataset), and provides the best performing model.


There seems to be no strong correlation between the number of parameters and the variance of a model performance with respect to the random initialization of its parameters. This is surprising, given the intuition that more parameters can potentially lead to a lower probability of being stuck in a local minima. The case may be that different initializations lead to different training times required to get to good local minimas.


\section{Conclusion}
One of the main outcomes of our experiments is that sequence-level optimization is key to achieve the best performance. Moreover, augmenting any neural architecture with a CRF layer on top has a very low cost in terms of parameters and a very good return in terms of performance. Our best performing models (in terms of average F$_1$) are LSTM-CRF and LSTM-CRF-CHAR-REP. In general we may say that adding a sequence level control to different type of NN architectures leads to very good model performances.
Another important observation is the variance of performance of NN models with respect to initialization parameters. Consequently, we strongly believe that this variability should be taken into consideration and reported (with the lowest and highest performances) to improve the reliability and replicability of the published results.


\bibliographystyle{acl}
\bibliography{clic}

\end{document}